\documentclass[letterpaper, 10 pt, conference]{ieeeconf}  

\IEEEoverridecommandlockouts                              

\usepackage{eso-pic}
\usepackage{geometry}
\usepackage{caption} 
\usepackage{booktabs}
\usepackage{multirow}
\usepackage{makecell}
\usepackage{amssymb}
\usepackage{graphicx}
\usepackage{kotex}
\usepackage{subcaption}
\usepackage{xcolor}
\usepackage{amsmath}
\usepackage{algorithm}
\usepackage{algorithmic}

\title{\LARGE \bf
SPARSE RECOVERY BIRD’S-EYE VIEW MODEL SOUP FOR ROBUST
CAMERA-BASED 3D OBJECT DETECTION
}

\author{Jinwoo Chung$^{1}$, Sangho An$^{1}$, Hyunsung Park$^{2}$, and Jangho Kim$^{1 *}$
\thanks{This work was supported by Hyundai Motor Company and Kia, and partially supported by the Institute of Information \& Communications Technology Planning \& Evaluation(IITP) grant funded by the Korea government(MSIT) (No.RS-2025-02219317, AI Star Fellowship(Kookmin University)), the National Research Foundation of Korea(NRF) grant funded by the Korea government(MSIT) (RS-2025-23524783) and Institute of Information \& communications Technology Planning \& Evaluation(IITP) under the Leading Generative AI Human Resources Development(IITP-2024-RS-2024-00397085) grant funded by the Korea government(MSIT).}
\thanks{$^{1}$Kookmin University, Seoul, Republic of Korea, $^{2}$Hyundai Motor Company, $^{*}$Corresponding author: Jangho Kim}
}

\begin{document}

\maketitle
\thispagestyle{empty}
\pagestyle{empty}
\AddToShipoutPictureFG*{%
  \AtPageLowerLeft{%
    \put(\LenToUnit{0.75in},\LenToUnit{0.2in}){%
      \parbox[b]{\textwidth}{%
        \normalfont\fontsize{7}{8}\selectfont\raggedright
        \textcopyright~2026 IEEE.
        Personal use of this material is permitted. Permission from IEEE must be obtained for all other uses, in any current or future media, including reprinting/republishing this material for advertising or promotional purposes, creating new collective works, for resale or redistribution to servers or lists, or reuse of any copyrighted component of this work in other works.

      }%
    }%
  }%
}

\begin{abstract}
Camera-based bird's-eye view (BEV) perception methods such as Fast-BEV have demonstrated promising results with fast inference speed and competitive accuracy. However, their robustness to real-world corruptions remains insufficiently explored, posing risks for autonomous driving deployment where environmental variations are inevitable. In this paper, we propose \textit{Sparse Recovery Bird's-Eye View Model Soup (SR-BEVS)}, a sparse recovery learning approach that simultaneously improves clean accuracy and corruption robustness without requiring additional data or architectural modifications. Our method identifies pruning candidates based on layer-wise weight magnitudes, applies random pruning to create diverse sparse models, and retrains each model with a randomly sampled learning rate. By averaging these diverse recovered models into a single dense model, we achieve enhanced robustness against various corruptions while introducing zero inference overhead. We further provide a mode connectivity analysis demonstrating that our pruning-based diversification preserves the connected low-loss basin necessary for stable weight averaging. Experiments on the nuScenes dataset demonstrate up to 33.8\% improvement in corruption robustness and a 1.8\% gain in clean accuracy over the Fast-BEV baseline, with consistent improvements across sensor noise, weather distortions, and illumination changes. Our simple yet effective approach may be applicable to other BEV models.
\end{abstract}


\section{INTRODUCTION}

\label{sec:intro}

Camera-based Bird's-Eye View (BEV) perception offers a cost-effective approach for autonomous driving. Fast-BEV~\cite{fastbev} achieves real-time inference through efficient view transformation and a streamlined architecture, enabling deployment on embedded platforms with competitive detection accuracy. Importantly, this efficiency stems from architectural design rather than simply scaling up model size, avoiding the higher latency, memory footprint, and power consumption that would otherwise hinder practical deployment.

\begin{figure}[tb]
\centering
\includegraphics[width=0.9\linewidth]{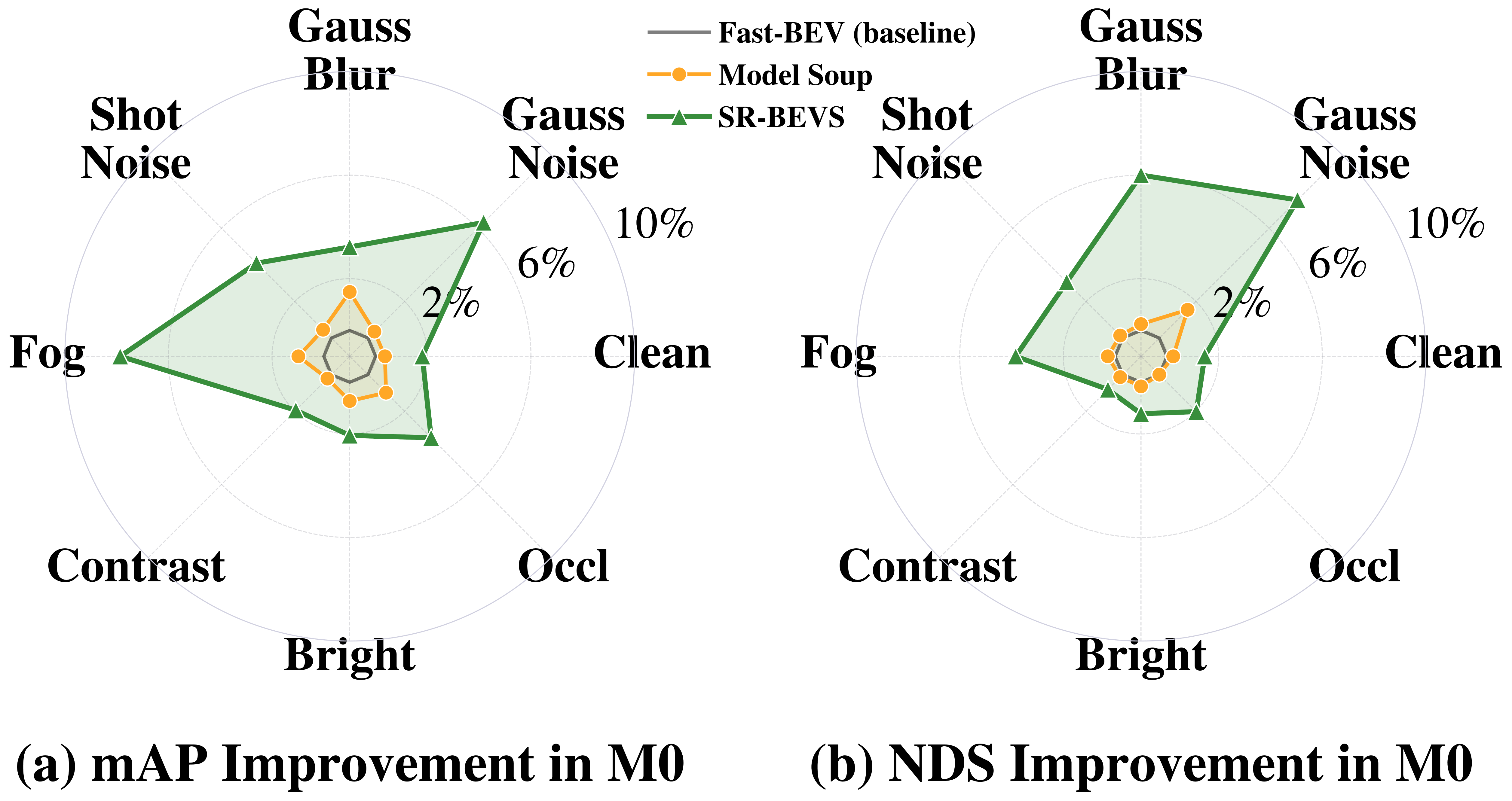}
\caption{The mAP (a) and NDS (b) performance improvement rates (\%) of Model Soup and 
SR-BEVS compared to the Fast-BEV baseline on the M0 model~\cite{fastbev}. Results are averaged across severity levels for each of the seven corruption types, plus the nuScenes clean data~\cite{nuscenes}. The results show that our SR-BEVS 
consistently outperforms the two baselines.
}
\label{fig1}
\end{figure}

\begin{figure*}[tb]
    \centering
    \includegraphics[width=0.98\linewidth]{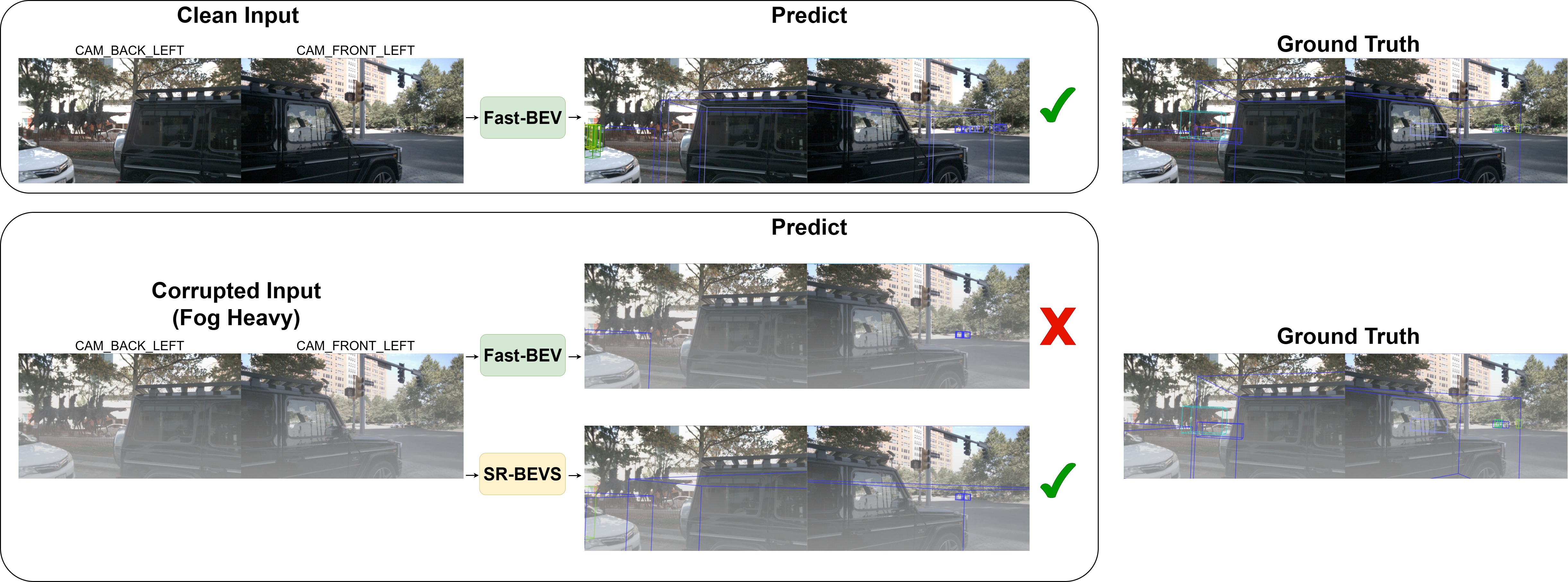}
\caption{Qualitative comparison under clean and corrupted inputs. Fast-BEV performs well on clean data but produces incorrect predictions under heavy fog, whereas SR-BEVS maintains accurate detection even under severe corruption, yielding more stable and spatially consistent bounding boxes across the scene.}
\label{fig:intro_import}
\end{figure*}

However, robustness to real-world environmental variations is equally critical for safe autonomous driving. Perception failures under adverse weather, illumination changes, or sensor degradation can directly compromise safety-critical decision making. As shown in Fig.~\ref{fig1}, a model trained solely on clean data is highly vulnerable to diverse weather conditions and input noise, with substantial performance gaps across all corruption types. Fig.~\ref{fig:intro_import} further illustrates this vulnerability: although Fast-BEV achieves accurate detection on clean inputs, its performance degrades drastically under corrupted conditions such as heavy fog, with the mAP dropping from 0.277 to as low as 0.017 under severe Gaussian noise. Despite its importance, the robustness of Fast-BEV under corruption scenarios has not been systematically evaluated, posing potential risks for real-world deployment.

\begin{figure}
  \centering
  \begin{subfigure}[b]{0.48\linewidth}
    \centering
    \includegraphics[width=\linewidth]{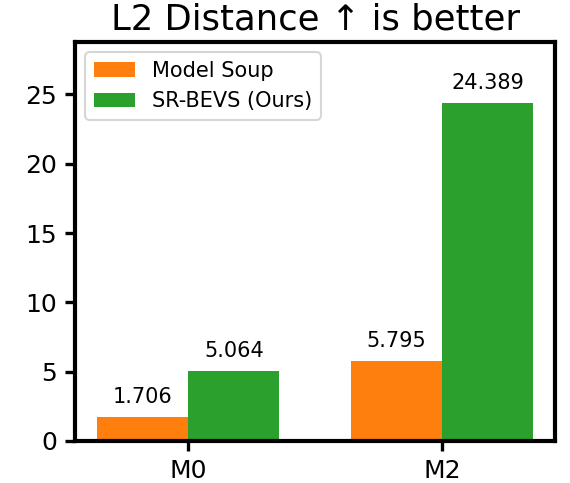}
    \caption{ \hspace{-2em} }
    \label{fig:pairwise_a}
  \end{subfigure}
  \hfill
  \begin{subfigure}[b]{0.48\linewidth}
    \centering
    \includegraphics[width=\linewidth]{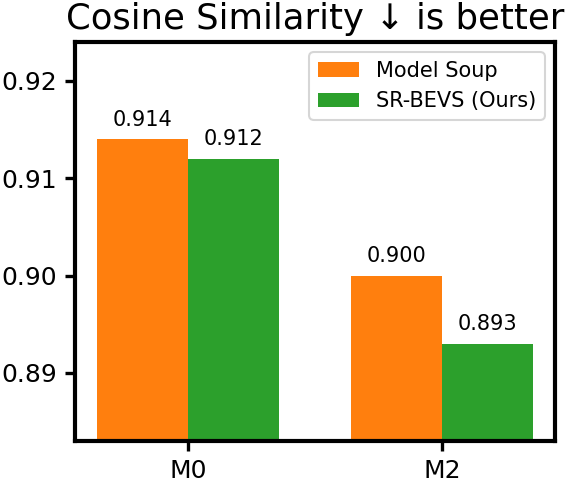}
    \caption{ }
    \label{fig:num_models}
  \end{subfigure}
\caption{Weight diversity comparison between Model Soup and SR-BEVS. (a)~L2 distance (higher is better) and (b)~cosine similarity (lower is better) between candidate model pairs. Standard Model Soup candidates exhibit high similarity, while SR-BEVS produces substantially more diverse models.}
  \label{fig:interp_combined}
\end{figure}

To address this robustness gap without increasing inference cost—a key constraint for autonomous driving—weight averaging of multiple fine-tuned checkpoints (model soup~\cite{modelsoup}) is appealing because it yields a single deployable model without architectural changes. However, applying it to Fast-BEV brings only marginal gains: due to the shared training recipe and strong camera-to-BEV inductive bias, independently fine-tuned checkpoints converge to highly similar solutions (Fig.~\ref{fig:interp_combined}), leaving little diversity for averaging to exploit. While one could force diversity via stronger perturbations or substantially different training dynamics, doing so risks leaving the same low-loss basin; once mode connectivity is broken, naive weight averaging/interpolation can cross high-loss regions and even hurt performance \cite{garipov2018loss,draxler2018essentially}. Thus, the core requirement is to increase diversity without breaking mode connectivity, enabling stable averaging with meaningful robustness gains.

We address this challenge by proposing \textit{Sparse Recovery Bird's-Eye View Model Soup (SR-BEVS)}, which \textbf{maintains mode connectivity within the same loss basin to enable stable averaging} while \textbf{deliberately increasing diversity} so that candidate models learn distinct representations. We provide a dedicated analysis of mode connectivity in Sec.~\ref{sec:method}, showing that SR-BEVS is carefully designed to inject diversity without leaving the connected low-loss region. Inspired by the weight averaging principle of model soup~\cite{modelsoup} and the well-known generalization benefits of connection removal in dropout~\cite{srivastava2014dropout}, SR-BEVS performs a \emph{one-time} random pruning step to induce distinct sparsity patterns across candidates, followed by a short recovery fine-tuning stage to restore accuracy.

Importantly, we prune \emph{low-magnitude} weights to avoid introducing excessive perturbations. This choice is primarily motivated by the long-standing observation in pruning literature~\cite{magni,magni2,magni3} that many small-magnitude parameters contribute little to the prediction function, so removing them can effectively preserve the model's overall performance while still creating controlled structural differences across candidates. By eliminating these relatively unimportant connections, we encourage complementary feature learning under different sparse patterns while keeping each candidate close to the original solution—thereby promoting diversity \emph{and} preserving mode connectivity. Finally, we average the recovered candidates into a single dense model, maintaining or improving clean accuracy while substantially reducing performance degradation under corruptions, with no inference overhead. As shown in Fig.~\ref{fig1}, SR-BEVS consistently outperforms both the Fast-BEV baseline and standard Model Soup across all corruption types, achieving up to 33.8\% improvement in corruption robustness and 1.8\% gain in clean accuracy.

Our contributions are summarized as follows:
\begin{itemize}
    \item We propose a simple yet effective sparse recovery learning method that improves both accuracy and robustness without additional inference cost. Our approach requires no architectural modifications and is validated on two Fast-BEV configurations, providing a practical training-time recipe for robustness improvement.

    \item Our method achieves significant improvements in both corruption robustness and clean accuracy over the Fast-BEV baseline, with an absolute NDS improvement from 0.077 to 0.103 under severe Gaussian noise (+33.8\% relative gain).

    \item We demonstrate that pruning-induced diversity, combined with mode connectivity within the same loss basin, enables effective weight averaging that goes beyond what standard model soup can achieve, providing practical insights for robust deployment.
\end{itemize}

\section{RELATED WORK}
\label{sec:format}
\noindent \textbf{Camera-based BEV Perception. }
LSS \cite{lss} pioneered 2D-to-3D transformation via depth estimation, followed by improvements like BEVDet \cite{bevdet}, BEVDepth \cite{bevdepth}, BEVFusion \cite{bevfusion} and Bevstereo \cite{bevstereo}. Transformer-based methods (BEVFormer \cite{bevformer}, DETR3D \cite{detr3d}) offer high accuracy but demand heavy computation. Fast-BEV \cite{fastbev}  uses pre-computed Look-Up-Tables to reach 52.6 FPS without depth prediction or transformers, balancing speed and accuracy for practical deployment. We select Fast-BEV as our baseline.

\noindent \textbf{Robustness in Autonomous Driving}
Autonomous vehicles encounter sensor noise, weather effects, and lighting changes in practice. ImageNet-C \cite{imagenet-c} standardized corruption testing for 2D images, but 3D detection from cameras lacks similar benchmarks. Current methods need data augmentation or adversarial training, yet cannot anticipate every real-world corruption. We enhance Fast-BEV using clean data alone, testing on 7 corruption types at 17 severity levels.

\noindent \textbf{Model Soup and Weight Averaging. }
Model soup~\cite{modelsoup} averages the weights of multiple fine-tuned models to improve accuracy and robustness without increasing inference cost. Prior work on loss surfaces~\cite{garipov2018loss, draxler2018essentially} showed that naive weight interpolation between models in different loss basins can cross high-loss regions and degrade performance, highlighting the importance of mode connectivity for stable averaging. A subsequent study on transfer learning~\cite{pretrained_same_basin} further demonstrated that models fine-tuned from the same pretrained initialization tend to remain in the same loss basin, enabling safe linear interpolation with low loss throughout the path. However, candidates differing only in learning rate or schedule tend to converge to highly similar solutions, limiting the gains from averaging. Diverse weight averaging~\cite{diverse} addressed this limitation by showing that securing diversity alongside connectivity is critical for achieving meaningful improvements in out-of-distribution robustness.

\noindent \textbf{Model Pruning. }
Model pruning is a representative approach to model compression, aiming to accelerate edge deployment by reducing parameter count and/or improving inference efficiency~\cite{magni,magni2,kim2023finding}. Depending on the granularity, pruning is commonly categorized into unstructured pruning and structured pruning. In practice, real hardware speedups are more readily obtained with structured pruning (e.g., removing channels or filters), whereas aggressive structured pruning often leads to substantial accuracy degradation and is difficult to apply at high pruning ratios~\cite{an2025sparse,shim2024snp}. By contrast, unstructured pruning can remove a large fraction of weights with relatively modest accuracy loss, but its latency benefits typically require specialized sparse kernels or API support to translate sparsity into actual speedups~\cite{kim2025exploring,lin2020dynamic}. Among various pruning criteria, magnitude pruning is one of the most widely used: weights (or filters) with small absolute values are treated as less important and are preferentially removed. In this work, we adopt random unstructured pruning over low-magnitude weights not for compression, but to induce diversity across candidate models while keeping perturbations controlled. Pruning with a large ratio or removing influential parameters can significantly distort the optimization landscape, potentially breaking mode connectivity and harming accuracy. Our design instead deletes weights that are likely to have limited influence on predictions, thereby enabling different sparse patterns to encourage complementary representations while remaining within the same connected low-loss basin. As a result, we preserve mode connectivity for stable weight averaging and improve robustness without introducing any inference overhead.

\section{Method}
\label{sec:method}

\subsection{Backbone Architecture: Fast-BEV}

We adopt Fast-BEV~\cite{fastbev} as our baseline for its efficiency. Fast-BEV transforms $N=6$ camera images $\mathbf{I} \in \mathbb{R}^{H \times W \times 3}$ to BEV space through Fast-Ray transformation, which assumes uniform depth along camera rays. Unlike depth-based or transformer-based view transformations that require learned depth prediction or expensive attention operations, Fast-Ray relies entirely on fixed camera geometry, making it highly suitable for deployment on resource-constrained platforms.

The key components are: \\
(1) \textbf{Look-Up-Table (LUT). } pre-computes 3D-to-2D projection indices offline using fixed camera intrinsic and extrinsic parameters, eliminating runtime projection calculations. Since these parameters are constant once the perception system is built, the projection index is computed only once and stored as a static table for efficient querying during inference. \\
(2) \textbf{Multi-View to One-Voxel. } directly projects all camera views to the same dense voxel feature, avoiding the expensive sparse voxel aggregation that would otherwise be needed when maintaining separate voxel features per camera. \\
(3) \textbf{Space-to-Channel (S2C). } operator reshapes 4D voxel tensor $\mathbf{V} \in \mathbb{R}^{X \times Y \times Z \times C}$ to 3D BEV tensor $\mathbf{V}' \in \mathbb{R}^{X \times Y \times (Z \cdot C)}$, avoiding expensive 3D convolutions.

We use single-scale image features with temporal fusion that aggregates 4 consecutive frames, resulting in 24 images (6 cameras $\times$ 4 timestamps) per sample. While Fast-BEV achieves competitive accuracy with fast inference, our evaluation reveals significant performance degradation under corruptions. We address this through our proposed SR-BEVS training strategy.

\begin{algorithm}[t]
\caption{SR-BEVS: Sparse Recovery Bird's-Eye View Model Soup}
\label{alg:srbevs}
\begin{algorithmic}[1]
\REQUIRE Pretrained Fast-BEV model $\theta_0$, pruning candidate ratio $p$, number of candidates $N$
\ENSURE Final robust model $\theta_{\text{final}}$

\STATE \textbf{// Stage 1: Candidate Pool Generation}
\FOR{each layer $l$ in $\theta_0$}
    \STATE Compute magnitude threshold $\tau_l$ at the $p$-th percentile of $|\mathbf{W}_l|$
    \STATE $\mathcal{C}_l \leftarrow \{w_i \in \mathbf{W}_l : |w_i| < \tau_l\}$
\ENDFOR

\STATE \textbf{// Stage 2: Pruning and Recovery}
\FOR{$k = 1$ \TO $N$}
    \STATE $\theta_k \leftarrow \theta_0$ \hfill // Clone pretrained weights
    \STATE Sample pruning ratio $p_k \sim \text{Uniform}(0, 1)$
    \STATE Sample learning rate $\alpha_k \sim \text{Uniform}(\alpha_{\min}, \alpha_{\max})$
    \FOR{each layer $l$}
        \STATE Randomly select $p_k$ fraction of weights from $\mathcal{C}_l$
        \STATE Set selected weights to zero in $\theta_k$
    \ENDFOR
    \STATE Fine-tune $\theta_k$ for $E$ epochs with learning rate $\alpha_k$ \hfill \\ // No sparsity mask
\ENDFOR

\STATE \textbf{// Stage 3: Weight Averaging}
\STATE $\theta_{\text{final}} \leftarrow \frac{1}{N} \sum_{k=1}^{N} \theta_k$ \hfill $\triangleright$ All $\theta_k$ are dense after recovery

\RETURN $\theta_{\text{final}}$
\end{algorithmic}
\end{algorithm}

\begin{figure}[tb]
    \centering
    \includegraphics[width=0.98\linewidth]{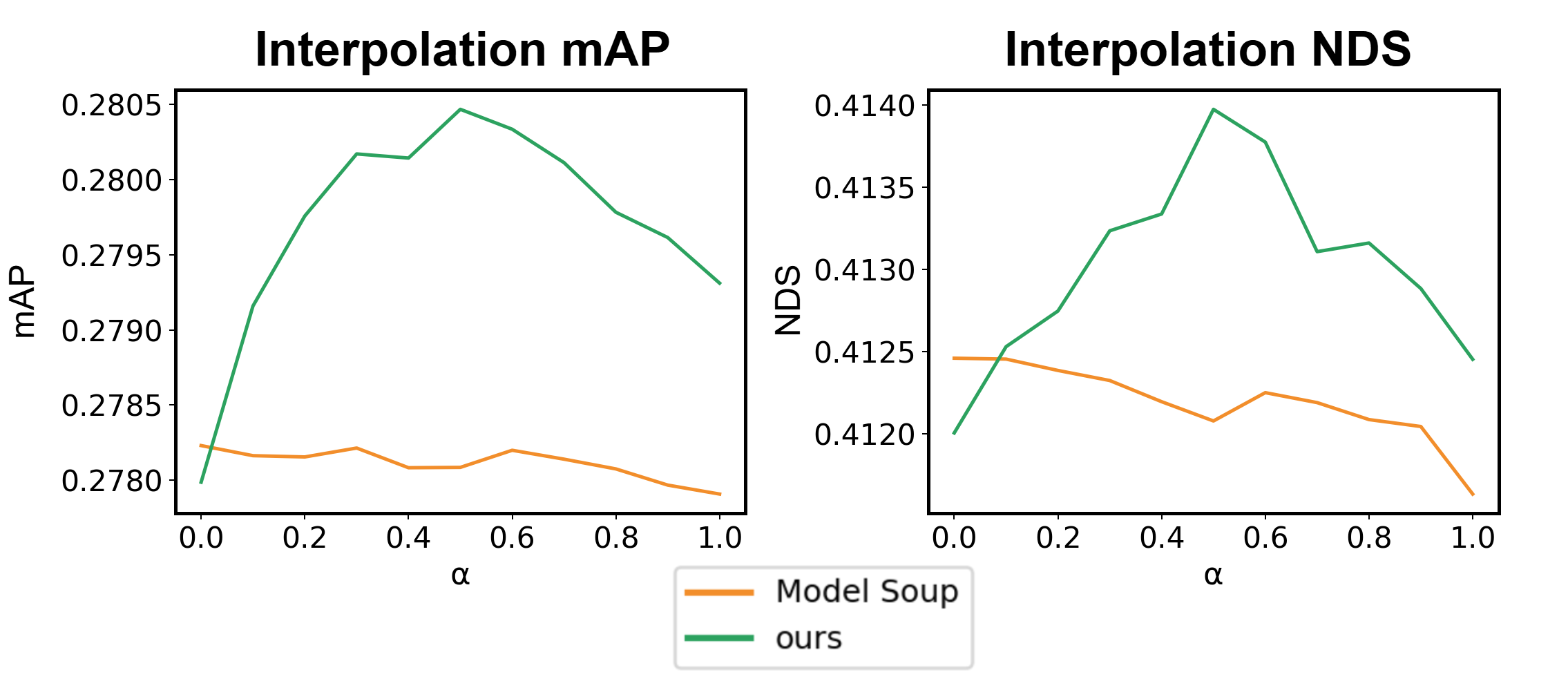}
\caption{Linear interpolation between SR-BEVS candidate model pairs on the M0 configuration. The horizontal axis denotes the interpolation coefficient $\alpha$ in $\theta(\alpha)=(1-\alpha)\theta_a+\alpha\theta_b$. Both mAP (left) and NDS (right) remain stable across the entire path, indicating the absence of a loss barrier between candidates.}
\label{fig:barrier}
\end{figure}

\subsection{Sparse Recovery Bird's-Eye View Model Soup}
\label{sec3.2}

To enhance both corruption robustness and clean accuracy, we propose a training strategy that creates diverse models through controlled pruning and recovery, then averages them into a single dense model for deployment. We adopt magnitude-based pruning principles, which have been shown to effectively identify less important weights~\cite{pruning_1}, preserve network capacity~\cite{pruning_2}, and maintain performance under various sparsity levels~\cite{pruning_3}.

\textbf{Stage 1: Candidate Pool Generation.}
For each layer $l$ in the pretrained Fast-BEV model $\theta_0$, we identify pruning candidates based on weight magnitudes:
\begin{equation}
\mathcal{C}_l = \{w_i \in \mathbf{W}_l : |w_i| < \tau_l\}
\end{equation}
where $\tau_l$ is the threshold chosen such that the pruning candidate ratio $p$ of the weights in layer $l$ (e.g., the lowest 20\% when $p=0.2$) fall below it. This layer-wise selection ensures that every layer contributes proportionally to the candidate pool, preventing any single layer from being excessively pruned. The pool $\mathcal{C}_l$ consists of the least important weights, whose removal introduces controlled perturbation while preserving the model's proximity to the original solution.

\textbf{Stage 2: Model Diversification via Pruning and Recovery.}
From the shared pretrained checkpoint $\theta_0$, we create $N$ diverse models through double randomization. For each model $k \in \{1, ..., N\}$, we:
\textbf{(1)} randomly sample a pruning ratio $p_k$;
\textbf{(2)} randomly select $p_k$ proportion of weights from the candidate pools $\{\mathcal{C}_l\}$ for pruning---this double randomization ensures each model has a unique sparsity pattern;
\textbf{(3)} prune the selected weights by setting them to zero, creating sparse model $\theta_k$;
\textbf{(4)} fine-tune $\theta_k$ with a randomly sampled learning rate $\alpha_k$.

Crucially, pruning is applied only once before training begins and no sparsity mask is enforced during fine-tuning. The pruned weights are free to recover non-zero values through gradient updates, so each candidate becomes a fully dense model after recovery. Because each candidate was forced to recover from a different damage pattern, the recovery process drives each model toward distinct feature representations and decision boundaries while remaining close to the original pretrained solution.

\textbf{Stage 3: Weight Averaging for Deployment.}
The final model is obtained through uniform parameter averaging:
\begin{equation}
\label{eq2}
\theta_{\text{final}} = \frac{1}{N} \sum_{k=1}^{N} \theta_k
\end{equation}
Since every candidate is a fully dense model after recovery, $\theta_{\text{final}}$ has exactly the same architecture and parameter count as the original Fast-BEV baseline, introducing zero inference overhead. An overview of our training pipeline is summarized in Algorithm~\ref{alg:srbevs}. When these diverse models with different learned pathways are averaged, their complementary strengths combine while individual vulnerabilities are mitigated, resulting in a model that is more robust to various corruptions. As we verify in Sec.~\ref{sec:loss_barrier}, this averaging is safe because all candidates remain within the same low-loss basin.

\begin{figure}[tb]
  \centering
  \begin{subfigure}[b]{0.49\linewidth}
    \centering
    \includegraphics[width=\linewidth]{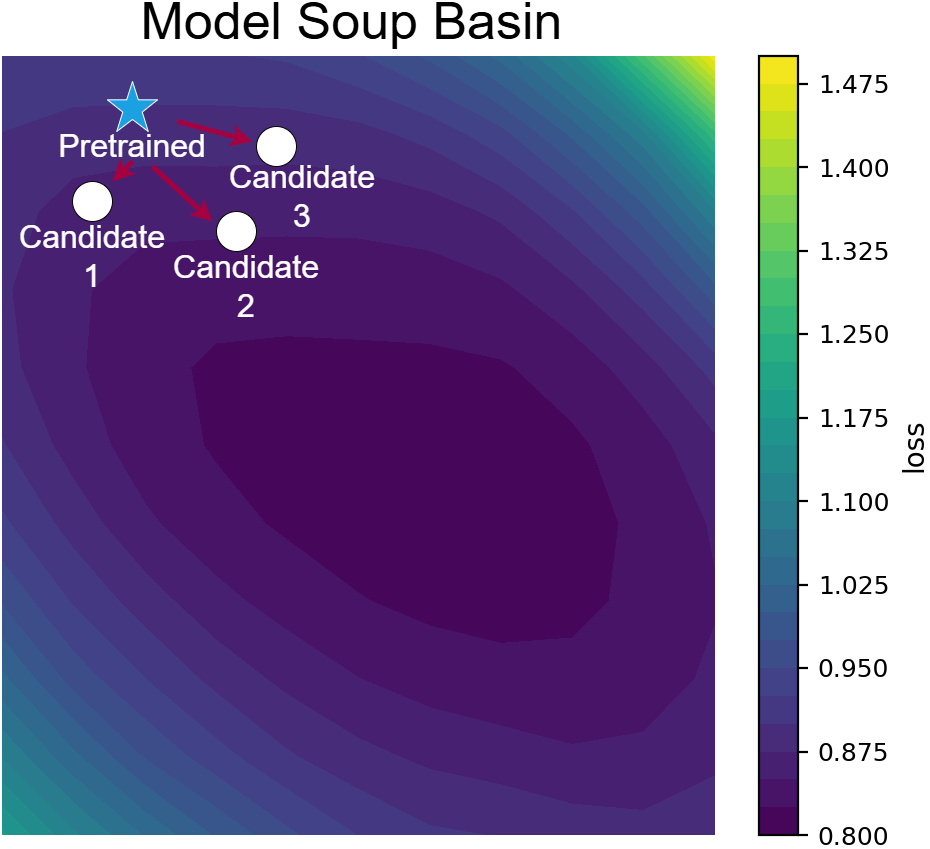}
    \caption{Model Soup}
    \label{fig:4a}
  \end{subfigure}
  \hfill
  \begin{subfigure}[b]{0.49\linewidth}
    \centering
    \includegraphics[width=\linewidth]{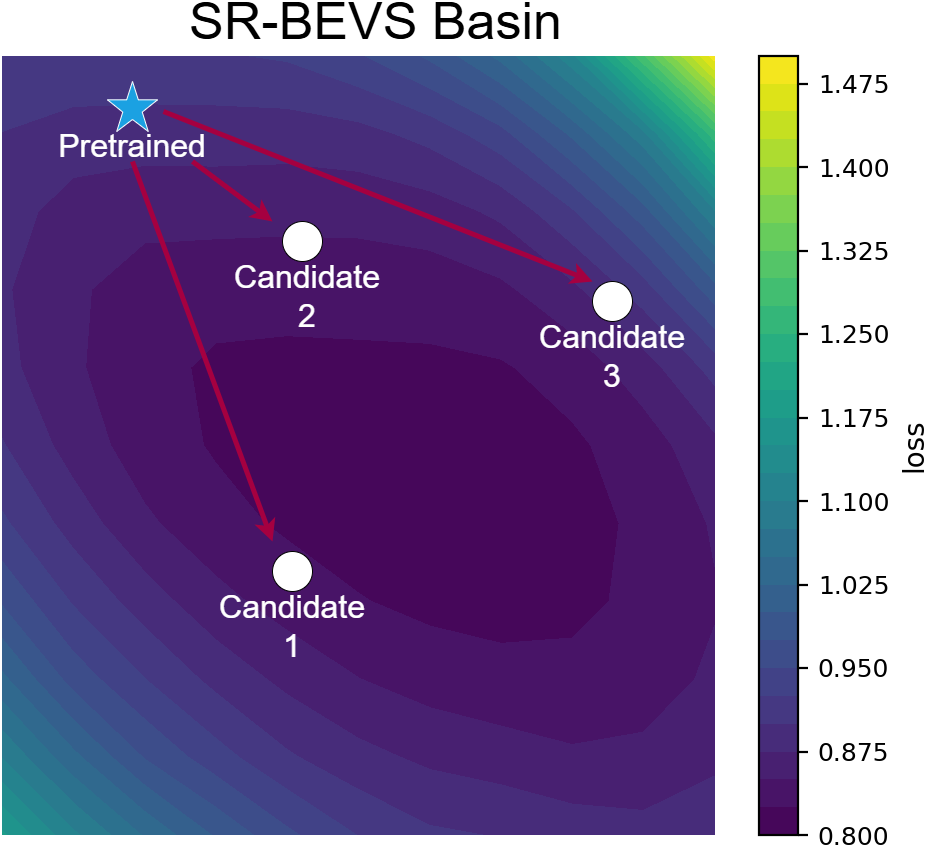}
    \caption{SR-BEVS (Ours)}
    \label{fig:4b}
  \end{subfigure}
\caption{2D loss landscape visualization on the M0 configuration. Candidate models are projected onto the two principal directions of the weight space. (a) Standard Model Soup candidates cluster tightly near the pretrained model. (b) SR-BEVS candidates are more spread out while remaining within the same low-loss basin, demonstrating increased diversity without breaking mode connectivity.}
  \label{fig:basin}
\end{figure}

\subsection{Mode Connectivity Analysis}
\label{sec:loss_barrier}

For weight averaging to improve performance, candidate models must reside in the same low-loss region or at least be connected through paths that do not cross high-loss areas. Prior work refers to this condition as \emph{mode connectivity}~\cite{garipov2018loss, draxler2018essentially} and has shown that models fine-tuned from the same pretrained initialization often satisfy it~\cite{pretrained_same_basin}. We verify this property for SR-BEVS both quantitatively and qualitatively through loss barrier analysis.

\paragraph{Loss barrier measurement protocol}
Given two candidate models $\theta_a$ and $\theta_b$, we construct a linear interpolation path $\theta(\alpha)=(1-\alpha)\theta_a+\alpha\theta_b$ for $\alpha\in[0,1]$ and evaluate performance on the validation set at each interpolation point. If performance collapses in the middle of the path, this indicates that the two models lie in different basins and averaging is unsafe. Conversely, if performance remains stable throughout, the models are connected within the same basin and weight averaging $\bar{\theta}=\frac{1}{K}\sum_k\theta_k$ can be expected to stay in the low-loss region.

\paragraph{Empirical verification}
Fig.~\ref{fig:barrier} shows the linear interpolation results between SR-BEVS candidate pairs on the M0 configuration. Both mAP and NDS remain stable across the entire interpolation path with no significant loss barrier observed. Furthermore, the 2D loss landscape visualizations in Fig.~\ref{fig:basin} confirm that SR-BEVS candidates are distributed within the same low-loss basin. In the standard Model Soup setting (Fig.~\ref{fig:4a}), candidates cluster tightly near the pretrained model due to limited diversity. In contrast, SR-BEVS candidates (Fig.~\ref{fig:4b}) are more spread out yet still remain within the same basin, indicating that our pruning-based diversification successfully increases inter-model distance without breaking mode connectivity.

\begin{table*}[t]
\centering
\caption{Robustness evaluation on nuScenes validation set under diverse corruptions and severity levels.
Fast-BEV (FT) uses the same additional fine-tuning budget as SR-BEVS recovery.}
\label{tab:robustness}
\setlength{\tabcolsep}{3.5pt}
\scriptsize
\begin{tabular}{l|l|l|ccc|ccc|cc|cc|cc|ccc|cc}
\toprule
& & \multirow{2}{*}{Method} & \multicolumn{3}{c|}{Gauss Noise} & \multicolumn{3}{c|}{Gauss Blur} & \multicolumn{2}{c|}{Shot Noise} & \multicolumn{2}{c|}{Fog} & \multicolumn{2}{c|}{Contr.} & \multicolumn{3}{c|}{Bright.} & \multicolumn{2}{c}{Occl.} \\
& & & 5\% & 10\% & 15\% & $\sigma$1 & $\sigma$2 & $\sigma$3 & 5\% & 10\% & M & H & L & H & 5\% & 10\% & 20\% & 10\% & 20\% \\
\midrule
\multirow{8}{*}{M0} 
& \multirow{4}{*}{mAP}
& Fast-BEV & 0.189 & 0.079 & 0.017 & 0.227 &  0.125 & 0.051 & 0.221 &0.219& 0.234 &  0.173 &  0.266 & 0.197 &0.276 &  0.276&0.275 & 0.226 &  0.180 \\
& & Fast-BEV (FT) & 0.190 & 0.081 & 0.017 & 0.227 &  0.126 & 0.052 & 0.222 &0.219& 0.234 &  0.174 &  0.267 & 0.198 &0.276 &  0.277&0.275 & 0.227 &  0.180 \\
& & Model Soup & 0.190 & 0.080 & 0.016 &0.229 & 0.127& 0.053 & 0.222 & 0.220 & 0.236 &0.175 &0.266 &   0.198 &0.278 & 0.278& 0.277 & 0.228 & 0.182  \\
& & \textbf{SR-BEVS} & \textbf{0.198} & \textbf{0.087} & \textbf{0.018} & \textbf{0.231} & \textbf{0.129} & \textbf{0.056} & \textbf{0.230} & \textbf{0.228} & \textbf{0.247} & \textbf{0.192} & \textbf{0.270} & \textbf{0.202} & \textbf{ 0.282} & \textbf{0.281} & \textbf{0.281} & \textbf{0.233} & \textbf{0.187} \\
\cmidrule{2-20}
& \multirow{4}{*}{NDS}
& Fast-BEV & 0.325 & 0.181 & 0.077 &  0.373 & 0.290 &0.153 &0.360 & 0.359 & 0.378 &0.325 & 0.403 & 0.339 &  0.412&0.411 &0.409& 0.369 & 0.324 \\
& & Fast-BEV (FT) & 0.326 & 0.181 & 0.080 &  0.374 & 0.290 &0.153 &0.360 & 0.360 & 0.378 &0.326 & 0.403 & 0.349 &  0.413&0.412 &0.409& 0.370 & 0.325 \\
& & Model Soup & 0.326 & 0.181 & 0.085 &0.375&0.291 & 0.152 & 0.360 & 0.360 & 0.378 &0.327 & 0.403  &  0.340 & 0.412 &0.412 &0.410 & 0.370 & 0.323  \\
& & \textbf{SR-BEVS} & \textbf{0.337} & \textbf{0.187} & \textbf{0.103} & \textbf{0.381} & \textbf{0.298} & \textbf{0.186} & \textbf{0.371} & \textbf{0.370} & \textbf{0.389} & \textbf{0.341} & \textbf{0.406} & \textbf{0.342} & \textbf{0.417} & \textbf{0.416} & \textbf{0.414} & \textbf{0.377} & \textbf{0.330} \\
\midrule
\multirow{8}{*}{M2}
& \multirow{4}{*}{mAP}
& Fast-BEV       & 0.218 & 0.104 & 0.031 & 0.275 & 0.155 & 0.054 &0.254 & 0.252 & 0.290 & 0.231 & 0.318 & 0.240& 0.325 & 0.324 &  0.322& 0.274 & 0.230 \\
& & Fast-BEV (FT)& 0.219 & 0.105 & 0.031 & 0.277 & 0.157 & 0.055 &0.256 & 0.254 & 0.291 & 0.231 & 0.319 & 0.240& 0.327 & 0.326 &  0.324& 0.276 & 0.232 \\
& & Model Soup   & 0.221 & 0.106 & 0.032 & 0.279 & 0.161 & 0.058 &0.259 & 0.256 & 0.293 & 0.233& 0.322 & 0.245 &0.330 & 0.330& 0.328 &  0.280 & 0.236 \\
& & \textbf{SR-BEVS} & \textbf{0.229} & \textbf{0.114} & \textbf{0.038} & \textbf{0.287} & \textbf{0.174} & \textbf{0.067} & \textbf{0.268} & \textbf{0.267} & \textbf{0.301} & \textbf{0.248} & \textbf{0.327} & \textbf{0.255} & \textbf{0.335} & \textbf{0.335} & \textbf{0.333} & \textbf{0.284} & \textbf{0.238} \\
\cmidrule{2-20}
& \multirow{4}{*}{NDS}
& Fast-BEV        & 0.370 & 0.228 & 0.151 & 0.416 & 0.321 & 0.171 & 0.403& 0.399 & 0.426 & 0.380& 0.450 & 0.381 & 0.453 & 0.453 & 0.450 & 0.414 & 0.374\\
& & Fast-BEV (FT)& 0.372 & 0.230 & 0.152 & 0.417 & 0.323 & 0.172 & 0.405& 0.401 & 0.427 & 0.381& 0.450 & 0.382 & 0.455 & 0.455 & 0.452 & 0.416 & 0.376\\
& & Model Soup   & 0.375 & 0.232 & 0.153 & 0.419 & 0.325 & 0.191 &  0.408 & 0.404& 0.432 & 0.384& 0.455 & 0.388 &0.459 & 0.459 &0.457 & 0.418  & 0.382\\
& & \textbf{SR-BEVS} & \textbf{0.383} & \textbf{0.254} & \textbf{0.161} & \textbf{0.429} & \textbf{0.342} & \textbf{0.203} & \textbf{0.413} & \textbf{0.411} & \textbf{0.435} & \textbf{0.391} & \textbf{0.460} & \textbf{0.396} & \textbf{0.462} & \textbf{0.461} & \textbf{0.460} & \textbf{0.423} & \textbf{0.385} \\
\bottomrule
\end{tabular}
\vspace{-2mm}
\label{tab:table_2}
\end{table*}

\paragraph{Implications for averaging gain}
The low loss barrier confirmed above is a \emph{necessary condition} that ensures weight averaging does not drift into high-loss regions. However, connectivity alone is not sufficient for meaningful improvement. When candidates are too similar, as in standard model soup where only learning rates differ, averaging provides limited gain. SR-BEVS satisfies both conditions simultaneously: (i) mode connectivity ensures safe averaging (Fig.~\ref{fig:barrier}, Fig.~\ref{fig:basin}), and (ii) pruning-induced diversity ensures that averaging is beneficial by combining complementary feature representations (Fig.~\ref{fig:interp_combined}). This dual satisfaction of connectivity and diversity explains the consistent improvements of SR-BEVS over both the baseline and standard model soup.

\section{Experiments}
\label{sec:typestyle}

\subsection{Experimental Setup}

\textbf{Dataset.}
We conduct experiments on the nuScenes dataset \cite{nuscenes}, utilizing only the camera modality for 3D object detection. The
dataset provides synchronized images from 6 surround-view cameras covering 360° around the vehicle: three front-facing cameras (CAM FRONT,
CAM FRONT LEFT, CAM FRONT RIGHT) and three rear-facing cameras (CAM BACK, CAM BACK LEFT, CAM BACK RIGHT). Each camera captures RGB
images at 1600 $\times$ 900 resolution. We use the standard split with 28,130 training samples (700 scenes) and 6,019 validation samples (150
scenes). For temporal context, we aggregate 4 consecutive frames at 3-frame intervals, resulting in 24 images (6 cameras $\times$ 4
timestamps) per sample. The dataset provides 3D bounding box annotations for 10 categories within a 100 m $\times$ 100 m BEV range.

\textbf{Implementation Details.}
We implement our method based on Fast-BEV with ResNet \cite{resnet} backbones (R18 for M0, R34 for M2). Following the Fast-BEV model family design, we evaluate two model configurations: M0 (R18 backbone) and M2 (R34 backbone) from their proposed M0-M5 series, which offer a good balance between efficiency and performance for our pruning experiments. All models are trained on 3 NVIDIA RTX 3090 GPUs. Input images are resized to 256 $\times$ 704. For our robust sparse ensemble, we set the pruning candidate ratio to $p=0.2$, identifying the bottom 20\% of weights by magnitude in each layer as pruning candidates, from which random subsets are selected to create diverse sparsity patterns. We create N=3 models using the baseline models. Learning rates are uniformly sampled from [5e-6, 5e-4] for each model. Models are fine-tuned for 2 epochs (M0) or 4 epochs (M2) using AdamW \cite{adamw} optimizer with batch sizes of 16 (M0) or 12 (M2) samples per GPU.

\textbf{Corruption Evaluation.} 
To systematically evaluate robustness in 3D detection, we designed a new corruption benchmark. This benchmark is inspired by the widely used ImageNet-C \cite{imagenet-c} for 2D images, but is specifically tailored to include corruptions commonly encountered in real-world driving scenarios. Specifically, we test on seven corruption types with varying severity levels, totaling 17 different corrupted validation sets:
(1) \textbf{Gaussian Noise} (5\%, 10\%, 15\% intensity): simulates sensor noise in low-light conditions;
(2) \textbf{Gaussian Blur} ($\sigma$=1, 2, 3): models camera focus issues and motion blur;
(3) \textbf{Shot Noise} (5\%, 10\% intensity): represents Poisson noise from camera sensors;
(4) \textbf{Fog} (mild, heavy): simulates adverse weather conditions;
(5) \textbf{Contrast} (low, high): captures varying lighting from overcast to bright sunlight;
(6) \textbf{Brightness} (5\%, 10\%, 20\% increase): models overexposure and bright environments;
(7) \textbf{Occlusion} (10\%, 20\% coverage): simulates partial view blockage from obstacles.

Each corruption is applied to all 6 camera views simultaneously to reflect realistic multi camera degradation. We evaluate on the full validation set (6,019 samples) for each corruption configuration.

\begin{table}[t]
\caption{Detection performance on nuScenes validation set. 
\textbf{Fast-BEV (FT)} denotes additional fine-tuning with the same training budget as SR-BEVS recovery.}
\centering
\begin{tabular}{l|cc|cc}
\hline
\multirow{2}{*}{Method} & \multicolumn{2}{c|}{M0 (R18)} & \multicolumn{2}{c}{M2 (R34)} \\
& mAP$\uparrow$ & NDS$\uparrow$ & mAP$\uparrow$ & NDS$\uparrow$ \\
\hline
Fast-BEV & 0.277 & 0.411 & 0.326 & 0.455 \\
Fast-BEV (FT) & 0.278 & 0.412 & 0.327 & 0.457 \\
Model Soup & 0.278 & 0.412 & 0.330 & 0.457 \\
\textbf{SR-BEVS} & \textbf{0.282} & \textbf{0.417} & \textbf{0.332} & \textbf{0.460} \\
\hline
\end{tabular}
\label{tab:table_1}
\end{table}

\textbf{Metrics.}
We evaluate performance using the nuScenes metrics: mAP (mean Average Precision) for detection accuracy and NDS (nuScenes Detection Score), a composite score that includes mAP and errors in key properties like translation, scale, and orientation to reflect detection quality.

\textbf{Baselines.}
We compare four methods that share the same architecture and parameter count at inference. Fast-BEV refers to the original pretrained model without any modification. Fast-BEV (FT) applies additional fine-tuning with the same training budget as our recovery stage to isolate the effect of extra training from that of pruning-induced diversity. Model Soup trains three models with different randomly sampled learning rates and averages their weights using~(\ref{eq2}). SR-BEVS averages three dense models recovered from differently pruned variants as detailed in Sec.~\ref{sec3.2}.

\textbf{Training and Inference Cost.}
Since the final model $\theta_{\text{final}}$ has the identical architecture and parameter count as a single Fast-BEV model, SR-BEVS introduces zero additional inference overhead in terms of latency, memory, and power consumption. The extra cost is limited to the training phase, where generating $N$ candidates requires $N$ independent recovery fine-tuning runs of $E$ epochs each. This cost scales linearly with $N$, but as shown in our ablation (Table~\ref{tab:table_num_models}), performance gains exhibit diminishing returns beyond $N{=}3$, making it a practical default choice.

\subsection{Quantitative Results}

\textbf{Clean Data Performance.}  Table~\ref{tab:table_1} shows that our method consistently outperforms both baselines across all configurations. For M0, we achieve 0.282 mAP and 0.417 NDS, surpassing Fast-BEV (0.277 mAP/0.411 NDS) and standard model soup (0.278 mAP/0.412 NDS). Similarly, M2 reaches 0.332 mAP and 0.460 NDS, exceeding baseline (0.326 mAP/0.455 NDS) and soup (0.330 mAP/0.457 NDS). While standard model soup provides marginal gains, our pruning-based approach delivers more substantial improvements, demonstrating that sparsity-induced diversity is more effective than simple fine-tuning diversity for model ensembling. \\

\begin{figure*}[tb]
    \centering
    \includegraphics[width=0.98\linewidth]{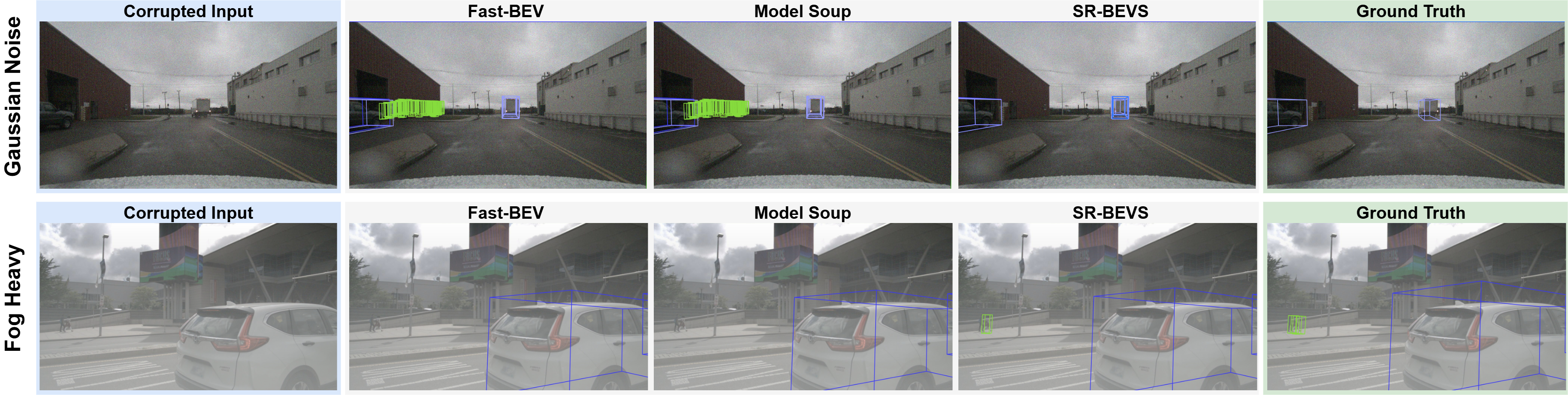}
\caption{Qualitative comparison of 3D detection results under severe corruptions on the M0 configuration. Top row: under strong Gaussian noise, Fast-BEV and Model Soup misclassify noise patterns as objects such as pedestrians (green boxes), while SR-BEVS avoids these incorrect predictions. Bottom row: under heavy fog, Fast-BEV and Model Soup miss objects present in the scene, whereas SR-BEVS maintains reliable detection consistent with the ground truth.}
\label{fig:qualitative}
\end{figure*}

\textbf{Corruption Robustness.} We evaluate robustness against critical corruptions commonly encountered in autonomous driving scenarios. We test seven corruption types that directly impact camera-based perception: sensor noise (Gaussian, Shot), motion artifacts (Blur), weather conditions (Fog), lighting variations (Contrast, Brightness), and partial occlusions. As shown in Table~\ref{tab:table_2}, our method consistently improves performance across all corruption types and severity levels for both M0 and M2 models. For M0, the most significant improvements occur under severe Gaussian noise (15\%: NDS 0.077→0.103, +33.8\%) and heavy fog (mAP 0.173→0.192, +11.0\%). M2 demonstrates even stronger gains with Gaussian blur ($\sigma=3$: NDS 0.171→0.203, +18.7\%) and heavy fog (mAP 0.231→0.248, +7.4\%). These consistent improvements across diverse corruptions demonstrate that SR-BEVS enhances model resilience to real-world visual degradations critical for safe autonomous driving deployment.

\subsection{Qualitative Results}
We select two corruption scenarios frequently encountered in real-world driving, heavy fog and severe Gaussian noise, and analyze the representative failure modes exhibited by each method.
\paragraph{Severe Gaussian noise: false positives}
Under strong Gaussian noise, the baseline tends to produce false positive detections by misinterpreting local high-frequency noise patterns as objects such as pedestrians in regions where no actual objects exist. This is a typical robustness failure where noise artifacts are mistaken for object cues. SR-BEVS substantially reduces these false positives under the same conditions. Weight averaging suppresses the extreme confidence scores and bounding box predictions of individual candidates, driving the ensemble toward more conservative and stable predictions.
\paragraph{Heavy fog: missed detections}
Under fog conditions, the overall contrast is severely degraded and object contours and texture cues become ambiguous. As a result, the baseline fails to detect certain objects that are present in the scene. SR-BEVS reliably detects these objects under the same conditions, maintaining a coherent BEV representation even in low-visibility environments. This can be attributed to the complementary predictions of the candidate models: when one candidate lacks sufficient cues to detect an object, other candidates that have learned alternative feature pathways can compensate, and their averaged prediction recovers the missed detection.
These qualitative observations suggest that SR-BEVS simultaneously mitigates the two dominant error modes under corrupted inputs: (i) missed detections caused by low visibility and (ii) false positives triggered by noise patterns. This is consistent with the mode connectivity analysis in Sec.~\ref{sec:loss_barrier}, where the low loss barrier enables safe averaging and the prediction complementarity across diverse candidates translates into tangible robustness gains.

\subsection{Ablation Study}

We conduct ablation experiments on the M0 configuration to analyze the impact of each design choice in SR-BEVS. Unless otherwise stated, all results are reported on the nuScenes validation set using clean data.

\textbf{Optimal Pruning Candidate Ratio.} Table~\ref{tab:table_4} investigates the impact of the pruning candidate ratio $p$ on model performance. We identify the bottom $p$\% weights by magnitude in each layer as pruning candidates, then randomly select subsets from this pool for actual pruning. A pruning candidate ratio of $p=0.2$ achieves the best balance with 0.282 mAP and 0.417 NDS, outperforming both $p=0.1$ (0.280 mAP, 0.414 NDS) and $p=0.3$ (0.279 mAP, 0.413 NDS). At $p=0.1$, the candidate pool is too small to generate sufficiently distinct sparsity patterns across candidates, limiting the diversity that drives averaging gains. At $p=0.3$, the pool includes weights of higher magnitude whose removal introduces excessive perturbation, making it difficult for models to fully recover within the short fine-tuning budget, and performance drops back toward the baseline. The $p=0.2$ setting strikes an effective balance between inducing enough structural perturbation to encourage diverse recovery pathways and preserving sufficient network capacity for the pruned weights to regain meaningful values during fine-tuning.

\begin{table}[t]
\centering
\caption{Effect of pruning candidate ratio $p$ on model performance (M0, clean data)}
\label{tab:pruning_ablation}
\begin{tabular}{ccc}
\toprule
\makecell{\textbf{Pruning Candidate}\\\textbf{Ratio ($p$)}} & \textbf{mAP} $\uparrow$ & \textbf{NDS} $\uparrow$ \\
\midrule
Baseline & 0.277 & 0.411 \\
$p=0.1$ & 0.280 & 0.414 \\
$p=0.2$ & \textbf{0.282} & \textbf{0.417} \\
$p=0.3$ & 0.279 & 0.413 \\
\bottomrule
\end{tabular}
\label{tab:table_4}
\end{table}

\begin{table}[t]
\centering
\caption{Effect of the number of candidate models $N$ on model performance (M0, clean data)}
\label{tab:num_models_ablation}
\begin{tabular}{ccc}
\toprule
\textbf{\#Candidates ($N$)} & \textbf{mAP} $\uparrow$ & \textbf{NDS} $\uparrow$ \\
\midrule
1 & 0.278 & 0.412 \\
2 & 0.280 & 0.414 \\
\textbf{3} & \textbf{0.282} & \textbf{0.417} \\
5 & 0.284 & 0.419 \\
\bottomrule
\end{tabular}
\label{tab:table_num_models}
\end{table}

\textbf{Number of Candidate Models.} Table~\ref{tab:table_num_models} examines how the number of candidate models $N$ affects the final averaged model. With $N=1$, the method reduces to a single pruned-and-recovered model without any averaging, which already slightly outperforms the Fast-BEV baseline due to the regularization effect of the pruning-recovery process. Increasing to $N=2$ brings an additional gain from weight averaging, and $N=3$ further improves performance to 0.282 mAP and 0.417 NDS. Extending to $N=5$ yields a marginal additional improvement of 0.284 mAP and 0.419 NDS, suggesting diminishing returns beyond three candidates. We use $N=3$ in our main experiments as it provides a favorable trade-off between training cost and performance, since each additional candidate requires a separate fine-tuning run.

\begin{table}[t]
\centering
\caption{Weight Diversity Analysis and Performance Comparison}
\resizebox{\columnwidth}{!}{%
\begin{tabular}{llcccccc}
\toprule
\multirow{2}{*}{Model} & \multirow{2}{*}{Method} & \multicolumn{2}{c}{Weight Diversity} & \multicolumn{2}{c}{Clean} & \multicolumn{2}{c}{Corrupted (Avg.)} \\
\cmidrule(lr){3-4} \cmidrule(lr){5-6} \cmidrule(lr){7-8}
 & & Cos.Sim.$\downarrow$ & L2 Dist.$\uparrow$ & mAP & NDS & mAP & NDS \\
\midrule
\multirow{2}{*}{M0} & Soup & 0.914 & 1.706 & 0.278 & 0.412 & 0.191 & 0.324 \\
     & \textbf{SR-BEVS} & \textbf{0.912} & \textbf{5.064} & \textbf{0.282} & \textbf{0.417} & \textbf{0.197} & \textbf{0.333} \\
\midrule
\multirow{2}{*}{M2} & Soup & 0.900 & 5.795 & 0.330 & 0.457 & 0.233 & 0.373 \\
     & \textbf{SR-BEVS} & \textbf{0.893} & \textbf{24.389} & \textbf{0.332} & \textbf{0.460} & \textbf{0.241} & \textbf{0.381} \\
\bottomrule
\end{tabular}%
}
\label{tab:table_3}
\end{table}

\textbf{Model Diversity Analysis.} We analyze the relationship between weight diversity and performance to validate our core hypothesis that pruning-induced diversity is the key driver behind the improvements of SR-BEVS. As shown in Table~\ref{tab:table_3}, SR-BEVS achieves substantially higher weight diversity compared to standard model soup across both model configurations. For M0, the L2 distance between candidate pairs increases approximately 3$\times$ from 1.706 to 5.064, while cosine similarity decreases from 0.914 to 0.912. The effect is even more pronounced for M2, where the L2 distance increases approximately 4$\times$ from 5.795 to 24.389 and cosine similarity drops from 0.900 to 0.893. This greater diversity in M2 is consistent with the larger model having more parameters available for pruning, which amplifies the structural differences created by distinct sparsity patterns.

Importantly, the enhanced diversity translates directly into performance gains, and the gains are more pronounced on corrupted data than on clean data. For M0, the corrupted-average NDS improves by 0.009 (0.324$\to$0.333) compared to a clean NDS improvement of 0.005 (0.412$\to$0.417). This pattern holds for M2 as well, with corrupted NDS improving by 0.008 versus 0.003 on clean data. This asymmetry suggests that diverse feature representations are particularly beneficial under distribution shift, where the complementary strengths of differently trained candidates provide broader coverage against varied corruption types. The consistent correlation between higher diversity and better robustness validates our hypothesis that sparse recovery learning creates ensembles whose diversity is both meaningful and beneficial.

\section{Conclusion}
\label{sec:majhead}
We investigated robustness limitations of camera-based BEV perception and found that Fast-BEV, while accurate on clean inputs, degrades sharply under realistic corruptions. To close this gap without increasing inference cost, we proposed \textit{Sparse Recovery Bird's-Eye View Model Soup (SR-BEVS)}, a training-time strategy that makes weight averaging effective by jointly preserving mode connectivity and injecting pruning-induced diversity. The recovered candidate models are averaged into a single dense network with identical architecture and zero inference overhead. Extensive experiments on nuScenes demonstrate that SR-BEVS consistently outperforms the Fast-BEV baseline and standard model soup, improving clean accuracy while substantially reducing performance degradation under corruptions (up to 33.8\% robustness gain under severe noise). While our validation is currently limited to Fast-BEV and synthetic corruptions, SR-BEVS offers a practical training-time recipe for improving robustness in deployment-constrained camera-based BEV perception.

\addtolength{\textheight}{-12cm}   






\bibliographystyle{IEEEtran}
\bibliography{bib}

\end{document}